\documentclass{article}

\usepackage{microtype}
\usepackage{subfigure}
\usepackage{nicefrac} 
\usepackage{hyperref}       
\usepackage{booktabs}       
\usepackage{bbm}
\usepackage{algorithm}
\usepackage{algorithmic}
\usepackage[table]{xcolor}

\usepackage{array}
\usepackage{amsthm,amsfonts,amsmath,amssymb,epsfig,color,float,graphicx,verbatim}
\graphicspath{ {./figures/} }
\usepackage[capitalize]{cleveref}

\newtheorem*{lemma*}{Lemma}

\newtheorem{assumption}{Assumption}
\newtheorem*{assumption*}{Assumption}

\usepackage{hyperref}

\usepackage[accepted]{icml2021}

\icmltitlerunning{Online Limited Memory Neural-Linear Bandits with Likelihood Matching}

\begin{document}

\twocolumn[
\icmltitle{Online Limited Memory Neural-Linear Bandits with Likelihood Matching}



\begin{icmlauthorlist}
\icmlauthor{Ofir Nabati}{to}
\icmlauthor{Tom Zahavy}{to,dm}
\icmlauthor{Shie Mannor}{to,nvidia}
\end{icmlauthorlist}

\icmlaffiliation{to}{Department of Electrical-Engineering, Technion Institute of Technology, Israel}
\icmlaffiliation{nvidia}{Nvidia Research}
\icmlaffiliation{dm}{DeepMind}
\icmlcorrespondingauthor{Ofir Nabati}{ofirnabati@gmail.com}

\icmlkeywords{Machine Learning, ICML}

\vskip 0.3in
]



\printAffiliationsAndNoticeArxiv{} 

\begin{abstract}
We study neural-linear bandits for solving problems where {\em both} exploration and representation learning play an important role. Neural-linear bandits harnesses the representation power of Deep Neural Networks (DNNs) and combines it with efficient exploration mechanisms by leveraging uncertainty estimation of the model, designed for linear contextual bandits on top of the last hidden layer.
In order to mitigate the problem of representation change during the process, new uncertainty estimations are computed using stored data from an unlimited buffer.
Nevertheless, when the amount of stored data is limited, a phenomenon called catastrophic forgetting emerges. To alleviate this, we propose a likelihood matching algorithm that is resilient to catastrophic forgetting and is completely online. 
We applied our algorithm, Limited Memory Neural-Linear with Likelihood Matching (NeuralLinear-LiM2) on a variety of datasets and observed that our algorithm achieves comparable performance to the unlimited memory approach while exhibits resilience to catastrophic forgetting.
\end{abstract}

\section{Introduction} 

Deep neural networks (DNNs) can learn complex representations of data and have dramatically improved the state-of-the-art in speech recognition, visual object recognition, object detection, and many other domains such as drug discovery and genomics \citep{lecun2015deep,goodfellow2016deep}. Using DNNs for function approximation in reinforcement learning (RL) enables the agent to generalize across states without domain-specific knowledge and learn rich domain representations from raw, high-dimensional inputs \citep{mnih2015human,SilverHuangEtAl16nature}.

Nevertheless, the question of how to perform efficient exploration during the representation learning phase is still an open problem. The $\epsilon$-greedy policy \citep{langford2008epoch} is simple to implement and widely used in practice \citep{mnih2015human}. However, it is statistically suboptimal. Optimism in the Face of Uncertainty (OFU) uses confidence sets to balance exploitation and exploration \citep{abbasi2011improved,auer2002using} while Thompson Sampling (TS) does so by choosing optimal action with respect to a sampled belief of the model.  \citep{thompson1933likelihood,agrawal2013thompson}. For DNNs, such confidence sets may not be accurate enough to allow efficient exploration. For example, using dropout as a posterior approximation for exploration does not concentrate on observed data \citep{Osband2018} and was shown empirically to be insufficient \citep{riquelme2018deep}. Alternatively, pseudo-counts, a generalization of the number of visits, were used as an exploration bonus \citep{bellemare2016unifying,pathak2017curiosity}. Inspired by tabular RL, these ideas ignore the uncertainty in the value function approximation in each context. As a result, they may lead to inefficient confidence sets \citep{Osband2018}.


Linear models, on the other hand, are considered more stable and provide accurate uncertainty estimates but require substantial feature engineering to achieve good results. Additionally, they are known to work in practice only with "medium-sized" inputs (with around $1,000$ features) due to numerical issues. 
A natural attempt at getting the best of both worlds is to learn a linear exploration policy on top of the last hidden layer of a DNN, which we term the \textbf{neural-linear} approach. In RL, this approach was shown to refine the performance of DQNs \citep{levine2017shallow} and improve exploration when combined with TS \citep{azizzadenesheli2018efficient} and OFU \citep{o2017uncertainty,zahavy2018learn}. For contextual bandits, \citeauthor{riquelme2018deep} (\citeyear{riquelme2018deep}) showed that neural-linear TS achieves superior performance on multiple datasets.

A practical challenge for neural-linear bandits is that the representation (the activations of the last hidden layer) changes after every optimization step, while linear contextual bandits algorithms 
assume the features are stationary. 
A recent line of work \cite{zhou2020neural, zhang2020neural, xu2020neural, domingos2020every} analyze deep contextual bandits in the infinite width regime, in which networks behave as a linear kernel method. This kernel is known as the Neural Tangent Kernel (NTK) \cite{jacot2018neural}.
Under the NTK assumptions, the optimal solution (and its features) are guaranteed to be close to the initialization point, which implies that the representation barely changes during training when neural bandits scheme is used. Under these assumptions, the authors of \citet{zhou2020neural} showed that the regret of a neural bandit is not larger than $\Tilde{O}(\sqrt{T})$ at round $T$ with a high probability. 

\citeauthor{riquelme2018deep} (\citeyear{riquelme2018deep}), on the other hand, observed that with standard DNN architectures, these features change during training (representation drift), and a mechanism to adapt for that change is required. They tackled this problem by storing the entire dataset in a memory buffer and computing new features for all the data after each DNN learning phase. The authors also experimented with a bounded memory buffer but observed a significant decrease in performance due to \textbf{catastrophic forgetting} \citep{Kirkpatrick3521}, i.e., a loss of information from previous experience.

In this work, we propose a complementary approach that can be added on top of the previously mentioned algorithms to tackle the representation drift problem with small memory usage while being resilient to the catastrophic forgetting. The key to our approach is a novel method to compute priors whenever the DNN features change. Specifically, we adjust the moments of the reward estimation conditioned on new features to match the likelihood conditioned on old features. We achieve this by solving a semi-definite program \citep[SDP]{vandenberghe1996semidefinite} to approximate the covariance and using the weights of the last layer as a prior to the mean. This way, we narrow the representation drift between network updates while using limited memory. 

We call our algorithm Limited Memory Neural-Linear with Likelihood Matching (NeuralLinear-LiM2 or LiM2 in short). While LiM2 can be used for OFU based algorithms, we focus on the TS approach. 
To make LiM2 more appealing for real-time usage, we implement it in an online manner, in which updates of the DNN weights and the priors are done simultaneously every step by using stochastic gradient descent (SGD) followed by projection of the priors. This obviates the need to process the whole memory buffer after each DNN learning phase and keeps our algorithm's computational burden small.

We performed experiments on several real-world and simulated datasets,  using Multi-Layered Perceptrons (MLPs). These experiments suggest that our prior approximation scheme improves performance significantly when memory is limited and shows an advantage over the NTK based algorithms. We demonstrate that LiM2 performs well in a sentiment analysis dataset where the input is high dimensional natural language, and we use a Convolution Neural Network (CNN).

LiM2 is the first neural-linear algorithm that is resilient to catastrophic forgetting due to limited memory to the best of our knowledge. In addition, unlike \citet{riquelme2018deep}, which uses multiple iterations per learning phase, LiM2 can be configured to work in an online manner, in which the DNN and statistics are efficiently updated each step. A code of our algorithm is based on the code provided by \citet{riquelme2018deep} and is available online \footnote{Code is available at  \href{https://github.com/ofirnabati/Neural-Linear-Bandits-with-Likelihood-Matching}{GitHub}}.

\begin{figure}
\centering
 \includegraphics[width=0.95\linewidth]{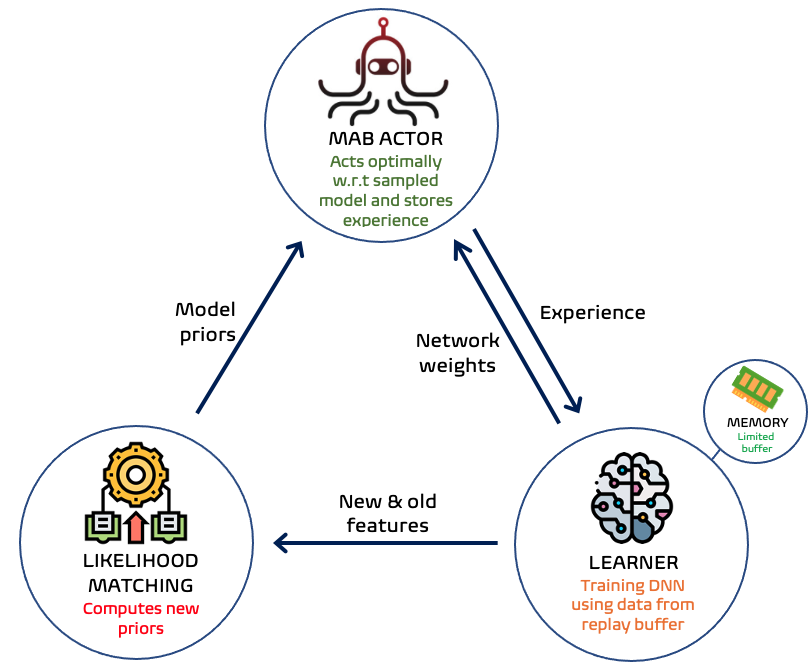}
 \caption{Scheme of our proposed method NeuralLinear-LiM2.}
 \label{fig:scheme}
\end{figure}

\section{Background} 

\textbf{The stochastic, contextual (linear) multi-armed bandit problem. }
At every round $t=1,2,3,..,T$, a contextual bandit algorithm observes a context $b(t)$ and chooses an arm $a(t) \in [1, \ldots, N]$. The bandit can use the history $H_{t-1}$ to make its decisions, where $H_{t-1} = \{b(\tau), a(\tau ), r_{a(\tau)}(\tau),  \tau =1, . . . , t - 1\},$  and $a(\tau)$ denotes the arm played at time $\tau$. 

Most existing works typically make the following \textbf{realizability} assumption \citep{chu2011contextual,abbasi2011improved,agrawal2013thompson}. 
\begin{assumption}
\label{ass:realizablity}
The reward for arm $i$ at time $t$ is generated from an (unknown) distribution s.t.
$\mathbb{E}\left[ r_i(t) | b(t),H_{t-1}\right] = \mathbb{E} \left[r_i(t)| b(t)\right] = b(t)^\top \mu_i,$
where $\{ \mu_i \in \mathbb{R}^d\}_{i=1}^N$ are fixed but unknown. 
\end{assumption}

Let $a^*(t)$ denote the optimal arm at time t, i.e. $a^*(t) = \text{arg max}_i b(t)^\top\mu _i,$ and let $\Delta_i(t)$ denote the difference between the mean rewards of the optimal arm and of arm $i$ at time $t$, i.e., $\Delta_i(t) = b(t)^\top \mu_{a^*(t)} - b(t)^\top\mu_{i}.$ The objective is to minimize the total regret $R(T) = \sum^T _{t=1} \Delta_{a(t)}$, where $T$ is finite.

        \begin{algorithm}[H]
        \caption{TS for linear contextual bandits}
        \begin{algorithmic}
        \STATE $\forall i \in [1,..,N],$ set $\Phi_i = 0$, $\Phi_i^0 = I_d$, $\hat{\mu}_i = 0_d$, $\psi_i = 0_d$
        \FOR{ $t = 1,2,\ldots,$ }   
            \STATE $\forall i \in [1,..,N],$ sample $\Tilde{\mu_i} \sim N (\hat{\mu_i}, \nu^2(\Phi_i^0 + \Phi_i)^{-1})$
            \STATE \textbf{Play} arm $a(t) := \text{argmax}_i  b(t)^\top \Tilde{\mu_i}$
            \STATE \textbf{Observe} reward $r_t$
            \STATE \textbf{Posterior update:} 
            \STATE \quad $\Phi_{a(t)} = \Phi_{a(t)} + b(t)b(t)^\top$
            \STATE \quad $\psi_{a(t)} = \psi_{a(t)} + b(t)r_t$
            \STATE \quad $\hat{\mu}_{a(t)} = (\Phi_i^0 + \Phi_{a(t)})^{-1}\psi_{a(t)}$
        \ENDFOR
        \end{algorithmic}
        \label{alg1}
        \end{algorithm}
\textbf{TS for linear contextual bandits.}
\label{sec:alg}
Thompson sampling is an algorithm for online decision problems where actions are taken sequentially in a manner that must balance between exploiting what is known to maximize immediate performance and investing to accumulate new information that may improve future performance \citep{russo2018tutorial,lattimore2018bandit}. For linear contextual bandits, TS was introduced by \citet{agrawal2013thompson} and is presented in Algorithm~\ref{alg1}. 

Suppose that the \textbf{likelihood} of reward $r_i(t),$ given context $b(t)$ and parameter $\mu_i$, was given by the pdf of Gaussian distribution
$Pr(r_i(t)| b(t), 
\mu_i)\propto \mathcal{N} (b(t)^\top \mu_i, \nu^2),$
and let $\Phi_i(t) =  \sum^{t-1}_{\tau=1} b(\tau)b(\tau)^\top 1_{i=a(\tau)},$ $\hat{\mu}_i(t) = (\Phi_i^0+\Phi_i(t))^{-1}\sum^{t-1}_{\tau=1} b(\tau)r_{a(\tau)}(\tau)1_{i=a(\tau)},$ 
where $1$ is the indicator function and $\Phi_i^0$ is the precision prior, initialized to $I_d$. Given a Gaussian \textbf{prior} for arm $i$ at time $t,$ $Pr(\Tilde{\mu}_i) \propto \mathcal {N}(\hat{\mu}_i(t), \nu^2(\Phi_i^0+\Phi_i(t))^{-1})$, the \textbf{posterior} distribution at time $t + 1$ is given by, 
\begin{align*}
    Pr(\Tilde{\mu}_i|r_i(t),b(t)) & \propto Pr(r_i(t)|b(t),\Tilde{\mu}_i) Pr(\Tilde{\mu}_i) \\ &\propto  \mathcal {N} (\hat{\mu}_i(t + 1), \nu^2(\Phi_i^0+\Phi_i(t+1))^{-1}).
\end{align*}

At each time step $t$, the algorithm generates samples $\{ \Tilde{\mu}_i(t)\}_{i=1}^N$ from the posterior distribution $\mathcal {N} (\hat{\mu}_i(t), \nu^2(\Phi_i^0 + \Phi_i(t))^{-1}),$ plays the arm $i$ that maximizes $b(t)^\top \tilde{\mu}_i(t)$ and updates the posterior.
TS is guaranteed to have a total regret at time $T$ that is not larger than $O (d^{3/2}\sqrt{T})$, which is within a factor of $\sqrt{d}$ of the information-theoretic lower bound for this problem. It is also known to achieve excellent empirical results \citep{lattimore2018bandit}. 
Although that TS often regarded as a Bayesian approach, the description of the algorithm and its analysis are prior-free, i.e., the regret bounds will hold irrespective of whether or not the actual reward distribution matches the Gaussian likelihood function used to derive this method \citep{agrawal2013thompson}.

\section{Our approach}
\textbf{Algorithm.}
LiM2 is composed of four main components: \textbf{(1)
Representation:} A DNN takes the raw context as an input and is trained to predict the reward of each arm; \textbf{(2) Exploration:} a mechanism that uses the last layer activations of the DNN as features and performs linear TS on top of them; \textbf{(3) Memory:} a buffer that stores previous experience; \textbf{(4) Likelihood matching:} a mechanism that uses the memory buffer and the DNN to account for changes in representation. Our full algorithm is presented in Algorithm~\ref{alg2}.

To derive LiM2, we make a \textbf{soft realizability assumption}, which is a stronger assumption than Assumption~\ref{ass:realizablity}. The difference is that we assume that \textbf{all} the representations produced by the DNN during training are \textbf{realizable} to some degree.


\begin{assumption}
\label{ass:realizablity2}
For any representation $\phi$ that is produced by the DNN during training with loss  \ref{eq:loss}, the reward for arm $i$ at time $t$ is generated from an (unknown) distribution s.t.
$\mathbb{E}\left[ r_i(t) | \phi(t),H_{t-1}\right] = \mathbb{E} \left[r_i(t)| \phi(t)\right]$ and $| \mathbb{E} \left[r_i(t)| \phi(t)\right] - \phi(t)^\top \mu_i| \leq \epsilon \quad \forall t,$
where $\{ \mu_i \in \mathbb{R}^d\}_{i=1}^N$ are fixed but unknown parameters and $\epsilon \geq 0$. 
\end{assumption}

That is, for each representation, there exists a \textit{different} linear coefficients vector such that the expected reward is approximately linear in the features. While this assumption may be too strong to hold in practice, it allows us to derive our algorithm as a good approximation that performs exceptionally well on many problems. 
We now explain how each of these components works.

\begin{algorithm}[ht]
\caption{Limited Memory Neural-Linear TS with Likelihood Matching (NeuralLinear-LiM2)}
\label{alg2}
\begin{algorithmic}
    \STATE \textbf{Set} $\forall i \in [1,..,N]: $ 
    \STATE \quad $\Phi^0_i = I_d, \hat \mu_i = \mu^0_i = 0_d, \Phi_i = 0_{dxd},\psi_i = 0_d $
    \STATE Initialize Replay Buffer $E$, and DNN $f$
    \STATE Define $\phi(t) \leftarrow \mbox{LastLayerActivations}(f(b(t)))$
\FOR{ $t = 1,2,\ldots,$ }
    \STATE \textbf{Observe} $b(t)$, evaluate $\phi(t)$
    \STATE $\forall i \in [1,..,N],$ sample $\Tilde{\mu_i} \sim N (\hat{\mu_i}, \nu^2(\Phi_i^0 + \Phi_i)^{-1})$
    \STATE \textbf{Play} arm $a(t) := \text{argmax}_i \phi(t)^\top \Tilde{\mu}_i$
    \STATE \textbf{Observe} reward $r_t$ and \textbf{Store} $\{b(t),a(t),r_t\}$ in $E$ 
    \IF{$E$ is full}
        \STATE Remove the first tuple in $E$ with $a=a(t)$ 
    \ENDIF
    \FOR{ $P$ steps} 
            \STATE \textbf{Sample} batch $\{b_j, a_j, r_j\}_{j=1}^{B}$ from $E$
            \STATE Compute old features $\{\phi_j^{old}\}_{j=1}^B$
            \STATE Optimize $\omega$ on $\nabla_{\omega} \mathcal{L}_{NN}$
            \STATE Compute new features $\{\phi_j\}_{j=1}^B$
            \STATE $(\Phi^0_i)^{-1} \leftarrow (\Phi_i^0 + \Phi_i)^{-1}$ 
            \FOR{ $\forall i \in [1,..,N]$ } 
                \STATE $e^{old}_i \leftarrow \{\phi_j^{old} | a_j=i\}, e_i \leftarrow \{\phi_j | a_j=i\} $ 
                \STATE $(\Phi^0_i)^{-1} \leftarrow                                      \mbox{PGD}((\Phi^0_i)^{-1}, e^{old}_i, e_i, \alpha)$
            \ENDFOR
        \ENDFOR    
        \FOR{ $\forall i \in [1,..,N]$ }
            \STATE Use the current weights of the last layer of the DNN as a prior for $\mu^0_i$             
            \STATE $\Phi_{i} = \sum_{j=1}^{n_i} \phi^i_j(\phi^i_j)^\top, \psi_i = \sum_{j=1}^{n_i}(\phi_j^i)^\top r_j.$ 
        \ENDFOR
        \STATE \textbf{Posterior update:} \par
        \STATE  \quad $\Phi_{a(t)} = \Phi_{a(t)} + \phi(t)\phi(t)^\top, \psi_{a(t)} = \psi_{a(t)}+ \phi(t)r_t,$ $\hat \mu_{a(t)} = (\Phi^0_{a(t)}+\Phi_{a(t)})^{-1}\left(\Phi^0_{a(t)} \mu^0_{a(t)} + \psi_{a(t)} \right) $\par

\ENDFOR
\end{algorithmic}
\end{algorithm}

\begin{algorithm}[ht]
\caption{ProjectedGradientDecent (PGD)}
\label{alg-pgd}
\begin{algorithmic}
\STATE \textbf{Inputs:} $A_0$ - PSD matrix , $\mathcal{B}_{old}, \mathcal{B}, \alpha$ 
\STATE Set $A \leftarrow A_0$
\FOR { $\phi_j^{old} \in  \mathcal{B}_{old}$ and $\phi_j \in \mathcal{B}$}
    \STATE \textbf{Gradient step:} \par
         \STATE \quad $s_j^2 \leftarrow  (\phi_j^{old})^\top A_0\phi_j^{old},$ \quad $X_j \leftarrow  \phi_j(\phi_j)^\top$ \par
         \STATE \quad $A \leftarrow A - \alpha \nabla_A(\text{Trace}(X_{j}^\top A) - s_{j}^2 )^2$ \par

    \STATE \textbf{Projection step:} \par
          \STATE \quad $A \leftarrow \mbox{EigenValueThresholding}(A)$ 
  
\ENDFOR
\end{algorithmic}
\end{algorithm}

\textbf{1. Representation.}
LiM2 uses a DNN, denoted by $f_{\omega}$, where $\omega$ represents the DNN's weights (for convenience, we omit the weight notation for the rest of the paper). The DNN takes the raw context $b(t)\in \mathbb{R}^d$ as its input. The network has $N$ outputs that correspond to the estimation of the reward of each arm, given context $b(t)\in \mathbb{R}^d,$ $f(b(t))_i$ denotes the estimation of the reward of the i-th arm. 

Using a DNN to predict each arm's reward allows our algorithm to learn a nonlinear representation of the context. This representation is later used for exploration by performing linear TS on top of the last hidden layer activations. We denote the activations of the last hidden layer of $f$ applied to this context as $\phi(t) = \text{LastLayerActivations}(f(b(t)))$, where $\phi(t)\in \mathbb{R}^g$. The context $b(t)$ represents raw measurements that can be high dimensional (e.g., image or text), where the size of $\phi(t)$ is a design parameter that we choose to be smaller ($g<d$). This makes contextual bandit algorithms practical for such datasets. 

\textbf{1.1 Training.}
Every iteration, we train $f$ for $P$ mini-batches. Training is performed by sampling experience tuples $\{b(\tau),a(\tau),r_{a(\tau)}(\tau)\}$ from the replay buffer $E$ (details below) and minimizing the mean squared error (MSE), \begin{equation}
\label{eq:loss}
    \mathcal{L}_{NN}(\omega) = ||f_\omega(b(\tau))_{a(\tau)} - r_{a(\tau)}(\tau)||^2_2,
\end{equation} 
where $r_{a(\tau)}$ is the reward that was received at time $\tau$ after playing arm $a(\tau)$ and observing context $b(\tau)$. Notice that only the output of arm $a(\tau)$ is differentiated, and that the DNN (including the last layer) is trained end-to-end to minimize Eq~\ref{eq:loss}.

\textbf{2. Exploration.}
Exploration is performed using the representation $\phi$. Similar to the linear case, we assume that, given $\phi$ and $\mu$, the likelihood of the reward is Gaussian. At each time step $t,$ the agent observes a raw context $b(t)$ and uses the DNN $f$ to produce a feature vector $\phi(t).$ The features $\phi(t)$ are used to perform linear TS, similar to Algorithm~\ref{alg1}, but 
instead of using a Gaussian posterior, we use the Bayesian Linear Regression (BLR) formulation that was suggested in \cite{riquelme2018deep}. Empirically, this update scheme was shown to converge to the true posterior and demonstrated excellent empirical performance \citep{riquelme2018deep}. 

In BLR, the noise parameter $\nu$ (Alg.~\ref{alg1}) is replaced with a prior belief that is being updated over time. The \textbf{prior} for arm $i$ at time $t$ is given by
$$Pr(\tilde \mu_i,\tilde \nu_i^2) = Pr(\tilde \nu_i^2)Pr (\tilde \mu_i|\tilde \nu_i^2),$$ 
where $Pr(\tilde \nu_i^2) \propto InvGamma(a_i(t),b_i(t)),$ is an inverse-gamma distribution and the conditional prior density, and $ Pr (\tilde \mu_i |\tilde \nu_i^2 ) \propto \mathcal {N}\left( \hat{\mu}_i(t),\tilde \nu_i^2 (\Phi_i^0 + \Phi_i(t))^{-1} \right).$ Combining this prior with a Gaussian likelihood guarantees that the the \textbf{posterior} distribution at time $\tau = t + 1$ is given in the same form (a conjugate prior). 

In each step and for each arm $i \in 1 .. N,$ we sample a noise parameter $\tilde \nu_i^2$ from $Pr(\tilde \nu_i^2)$ and then sample a weight vector $\tilde \mu_i$ from the posterior $\mathcal {N} \left(\hat{\mu}_i, \tilde \nu_i ^2(\Phi^0_i+\Phi_i)^{-1}\right).$ Once we sampled a weight vector for each arm, we choose to play arm $a(t) = \text{arg max}_i \phi(t)^ T \Tilde{\mu}_i, $ and observe reward $r_{a(t)}(t).$ This is followed by a posterior update step:

\begin{equation}
\begin{split}
\Phi_{a(t)} &=   \Phi_{a(t)} + \phi(t)\phi(t)^\top, \\ 
\psi_{a(t)} &= \psi_{a(t)}+ \phi(t)r_t , \nonumber \\
\hat \mu_{a(t)}  &= (\Phi_{a(t)}^0 + \Phi_{a(t)})^{-1}\big(\Phi^0_{a(t)} \mu^0_{a(t)} + \psi_{a(t)}\big), \\  
R^2_{i}(t)  &= R^2_{i}(t-1) + r_{i}^2, \\
A_{i}(t)  &= A^0_{a(t)}+{\frac {t}{2}}, \\  
B_{i}(t) &= B^0_{a(t)} + \frac{1}{2} \big( R^2_{i}(t) +({\mu }^0_{a(t)})^{\top} \Phi^0_{a(t)} \mu ^0_{a(t)}\\
&\quad -\hat{\mu}_{a(t)}^\top \Phi_{a(t)} \hat{\mu }_{a(t)} \big),
\end{split}
\end{equation}
where $\mu_i^0$ is the mean prior, initialized to $0_d$. We note that the exploration mechanism only chooses actions; it \textbf{does not change} the DNN's weights.


\textbf{3. Memory.}
After an action $a(t)$ is played at time $t,$ we store the experience tuple $\{b(t),a(t),r_{a(t)}(t)\}$ in a finite memory buffer of size $n$ that we denote by $E.$ Once $E$ is full, we remove tuples from $E$ in a FIFO manner, i.e., we remove the first tuple in $E$ with $a=a(t)$.

\textbf{4. Likelihood matching.}
After each training phase, we evaluate the features of $f$ on the replay buffer. Let $E_i$ be a subset of memory tuples in $E$ at which arm $i$ was played, and let $n_i$ be its size. We denote by $E^i_{\phi^{old}} \in \mathbb{R}^{n_i \times g}$ a matrix whose rows are feature vectors that were played by arm $i$. After a training phase is completed, we evaluate the new activations on the same replay buffer and denote the equivalent set by $E^i_\phi \in \mathbb{R}^{n_i \times g}$.

Our approach is to summarize the knowledge that the algorithm has gained from exploring with the old representation into the priors $\Phi^0_i,\mu^0_i$ under the new representation. Once these priors are computed, we restart the linear TS algorithm using the data that is currently available in the replay buffer. For each arm $i$, let $\phi^i_j = (E^i_{\phi})_j$ be the j-th row in $E^i_{\phi}$ and let $r_j$ be the corresponding reward, we set $\Phi_{i} = \sum_{j=1}^{n_i} \phi^i_j(\phi^i_j)^\top, \psi_i = \sum_{j=1}^{n_i}\phi_j^ir_j.$  We now explain how we compute $\Phi^0_i,\mu^0_i.$ 

Recall that under Assumption~\ref{ass:realizablity2}, the likelihood of the reward is approximately invariant to the choice of representation:
\begin{align*}
   \mathbb{E} [r_i(t)| \phi(t)] &\approx \phi(t)^\top \mu_i \\&\approx  \phi^{old}(t)^\top \mu ^{old}_i \approx \mathbb{E} [r_i(t)| \phi^{old}(t)]. 
\end{align*}

For all $i$, we define the estimator of the reward as $$\theta _i(t) = \phi(t)^\top \tilde \mu_i (t).$$ 
Due to the posterior distribution of $\tilde\mu_i (t)$, the marginal distribution of each $\theta_i(t)$ is $\mathcal {N}(\phi_i(t)^\top \hat \mu_i(t), \nu_i^2 s_{t,i}^2)$, where $s_{t,i}=\sqrt{\phi(t)^\top \Phi_i(t)^{-1}\phi(t)}$ (see \citet{agrawal2013thompson} for derivation).
The goal is to match the likelihood of the reward estimation $\theta _i(t)$ given the new features to be the same as with the old features. 

\textbf{4.1 Approximation of the mean} $\boldsymbol{\mu^0_i:}$ 
The DNN is trained to minimize the MSE (Eq~\ref{eq:loss}). Given the new features $\phi,$ the current weights of the last layer of the DNN already make a good prior for $\mu^0_i$. In \citet{levine2017shallow}, this approach was shown empirically to improve the performance of a neural-linear DQN. The main advantage is that the DNN is optimized online by observing the entire data and is therefore not limited to the current replay buffer. Thus, the weights of the current DNN hold information on more data and make a strong prior. 

\textbf{4.2 Approximation of the correlation matrix} $\boldsymbol{\Phi_{i}^0:}$ For each arm $i,$ LiM2 receives as input the sets of new and old features $E^i_{\phi},E^i_{\phi^{old}}$ with elements $\{\phi^{old}_j,\phi_j\}_{j=1}^{n_i}.$ In addition, the algorithm receives the correlation matrix $\Phi_i^{old}$. Notice that due to our algorithm's nature, $\Phi_i^{old}$ holds information on contexts that are not available in the replay buffer. The goal is to find a  correlation matrix,$\Phi^0_i$, for the new features that will have the same variance on past contexts as $\Phi_i^{old}.$ I.e., we want to find $\Phi^0_i$ such that 
\begin{equation*}
\begin{split}
    s_{j,i}^2  = \phi_j^\top(\Phi^0_i)^{-1}\phi_j
      = \text{Trace}\left((\Phi^0_i)^{-1}\phi_j\phi_j^\top\right) \\
      \forall i\in[1..N],j\in[1..n_i],
\end{split}
\end{equation*}

where $s_{j,i}^2  \doteq (\phi^{old}_j)^\top(\Phi_i^{old})^{-1} \phi^{old}_j$ and the last equality follows from the cyclic property of the trace. 
For hand $i \in [1..N]$, we denote by $X_{j,i} \doteq \phi_j\phi_j^\top \in \mathbb{R}^{d \times d}$ the 1-rank symmetric matrix $\forall j \in [1..n_i]$. Using this definition, we have that
$$
\text{Trace}\left((\Phi^0_i)^{-1}\phi_j\phi_j^\top\right) = \text{Trace}(X_{j,i}^\top(\Phi^0_i)^{-1}) 
$$
is an inner product over the vector space of symmetric matrices, known as the Frobenius inner product.
Finally, as $(\Phi^0_i)^{-1}$ is an inverse correlation matrix, we constrain the solution to be semi positive definite. Thus, the optimization problem is equivalent to a linear regression problem in the vector space of positive semi definite (PSD) matrices for all actions $i\in[1..N]:$
\begin{equation}
\begin{split}
\underset{(\Phi^0_i)^{-1}}{\text{minimize}}  \sum\nolimits_{j=1}^{n_i} \left(\text{Trace}(X_{j,i} ^\top(\Phi^0_i)^{-1}) - s_{j,i}^2 \right)^2\\  \text{subject to} \enspace \enspace (\Phi^0_i)^{-1} \succeq 0.
\label{sigma_opt}
\end{split}
\end{equation}
In practice, we solve the SDP by applying SGD using sampled batches from $E^i_{\phi^{old}}$ and $E^i_{\phi}$.  Each SGD iteration is followed by eigenvalues thresholding (denoted by $\mbox{EigenValueThresholding}((\Phi^0_i)^{-1})$) in order to project $(\Phi^0_i)^{-1}$ back to PSD matrices space. To avoid evaluating $E^i_{\phi}$ each time the DNN is updated, we take advantage of the iterative learning of the DNN and the iterative nature of the SGD by using the same batch to update the DNN weights and $(\Phi^0_i)^{-1}$ simultaneously. In each iteration, we treat the inverse correlation matrix from the previous iteration as $(\Phi_i^{old})^{-1}$ and also as the initial guess for the current gradient decent step. For each action $a \in A$, we use a subset of the batch, in which action $a$ was used.

\textbf{Computational complexity.}
We consider the time and memory complexity of LiM2 and their dependence on different parameters of the problem.
Recall that the last layer's dimension is $g<d$ where $d$ is the dimension of the raw features, the size of the limited memory buffer is $n$, the batch size is $B$, and the number of contexts seen by the agent is $T$. For the approximate SDP, each gradient step is $Bg^2$ (matrix-vector multiplications) plus the thresholding operator, which has a time complexity of $O(g^3)$ due to the matrix eigendecomposition. This can be improved in the case of low-rank stochastic gradients ($B<g$) into $O(g^2)$ as suggested in~\citet{chen2014efficient}. In addition, we use the contexts at our memory buffer to compute $\Phi_i$ after the prior matching, which is $O(n)$. The overall complexity is $O(T(g^2 + n))$ with a memory complexity of $O(n)$. 
On the other hand, the computational complexity of the full memory approach results is $O(T^2)$, and the memory complexity is $O(T)$. This is because it is estimating the TS posterior using the complete data every time the representation changes. Due to the stochastic behavior of our SGD modification, the computational complexity is linear in the batch size and not in $|A|$. 

To summarize, LiM2 is more efficient than the full memory baseline in problems with a lot of data ($T \gg g^2$ and $T \gg n$). Instead of solving an SDP after each update phase (which is computationally prohibitive in general), we apply an efficient SGD in parallel to the DNN updates. This is also sample efficient due to the reuse of the same batch for both tasks. In our experiments, we noticed that using only a single update iteration ($P=1$) for both the DNN and likelihood matching is enough to get competitive results. Therefore, this is our default configuration.   



\section{Experiments}
\label{sec:experiments}
In this section, we empirically investigate the performance of LiM2 to address the following questions:

     $\bullet$  Does the NTK assumption holds? (i.e., does representation drift occurs?)
     
     $\bullet$  Is our method sensitive to the memory buffer size?
     
     $\bullet$  Can neural-linear bandits explore efficiently while learning representations under finite memory constraints? 
     
     $\bullet$  Does the moment matching mechanism allows neural-linear bandits to avoid catastrophic forgetting? 
     
     $\bullet$  Can LiM2 be applied to a wide range of problems and across different DNN architectures?

We address these questions by performing experiments on ten real-world datasets, including high dimensional natural language data on a task of sentiment analysis from text, in which we evaluate LiM2 on a text-based dataset using CNNs. All of these datasets are publicly available through the UCI Machine Learning Repository \footnote{\url{https://archive.ics.uci.edu/ml/index.php}}. 

\begin{figure*}[ht]
\centering
    \includegraphics[trim=120 0 140 0, width=0.75\linewidth]{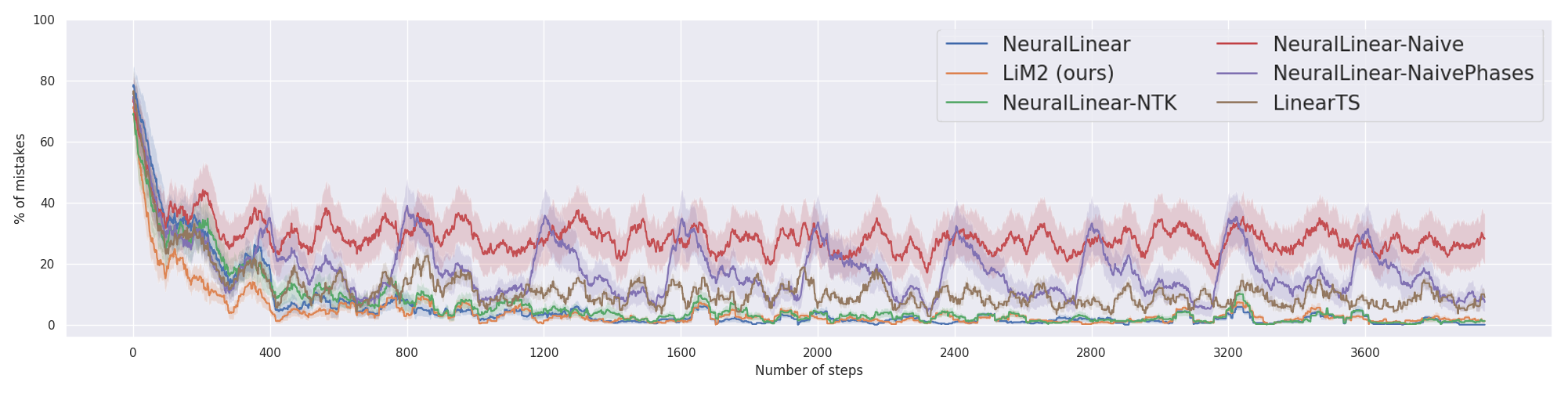}
    \label{fig:my_label}
\caption{\textbf{Catastrophic forgetting.} Limited memory based methods without likelihood matching suffer from performance degradation each training phase (\textcolor{purple}{purple} and \textcolor{red}{red}). Our method LiM2 (in \textcolor{orange}{orange}) performs similar to the unlimited memory version (in \textcolor{blue}{blue}).}
\label{fig:forgetting}
\end{figure*} 
 
\textbf{Methods and setup.} 
We experimented with different ablations of our approach, as well as a few baselines: 

\textbf{(1)} \textbf{LinearTS} \citep[Alg~\ref{alg1}]{agrawal2013thompson} using the raw context as a feature, with an additional uncertainty in the variance \citep{riquelme2018deep}. 

\textbf{(2) }\textbf{NeuralLinear} \citep{riquelme2018deep} with unlimited memory, in which the moments are computed each time the network is updated using all history. This strong baseline has shown to outperform other popular methods (Neural Greedy,
variational inference, expectation-propagation, dropout, Monte Carlo methods, bootstrapping, direct
noise injection, and Gaussian Processe. For more information regarding these methods see \citet{riquelme2018deep}) making it an important point of comparison.  

\textbf{(3)} \textbf{NeuralUCB} \cite{zhou2020neural} an NTK approach that uses UCB based exploration for a general reward function. This method chooses the action that maximizes the upper confidence bound (similar to \citet{li2010contextual}) but uses the network gradients as features and the network output as the mean reward estimator. The NTK assumptions claim that the network parameters stay very close to the initialization point under some conditions (most notably- the network width). Thus, it enables them to estimate the covariance matrix by accumulating the features' outer-product without recalculating them again. 

\textbf{(4)} \textbf{NeuralTS}  \citep{zhang2020neural} an NTK approach that uses TS based exploration for a general reward function. Like NeuralUCB, they use the estimated covariance matrix and the estimated mean to sample the reward for each arm and take the arm with the highest score. For more information regarding NeuralUCB and NeuralTS and their regret analysis, we refer the readers to \citet{zhang2020neural} and \citet{zhou2020neural}. Both NeuralUCB and NeuralTS implementations are based on the official code provided by the authors \footnote{\url{https://github.com/ZeroWeight/NeuralTS}}. 

\textbf{(5)} Our limited memory neural-linear TS algorithm with likelihood matching 
(\textbf{LiM2}).  

\textbf{(6)}  An ablative version of (5) that calculates the prior only for the mean, similar to \citet{levine2017shallow}. We call it Limited Memory Neural-Linear with Mean Matching (\textbf{NeuralLinear-MM}).

\textbf{(7)} An ablative version of (5) that does not use prior calculations. We call it Naive Limited Memory Neural-Linear (\textbf{NeuralLinear-Naive}). 

\textbf{(8)} Limited memory neural-linear TS with NTK assumption. This is the NTK version of LiM2 in which we do not perform likelihood matching (\textbf{NeuralLinear-NTK}) but still accumulate features without resetting the priors.

Algorithms 5-8 make an ablative analysis for LiM2. As we will see, adding each one of the priors improves learning and exploration. As noted, in all versions of LiM2, we update the network each round with one iteration, while in the full-memory version, we train the network every $400$ rounds with $800$ iterations due to the large computational burden that makes every round update prohibited. 
The full-memory NTK based algorithms (3-4) are trained every round with at most $1000$ iterations as their official code dictates. Note that in both cases, the overall number of training iterations is larger than LiM2.  
In all the experiments, we used the same hyperparameters as in \citet{riquelme2018deep}. E.g., the network architecture is an MLP with a single hidden layer of size $50$. The only exception is the text CNN (details below). The size of the \textbf{memory buffer} is set to be $100$ per action for the limited memory algorithms, and the batch size is set to be $16$ times the number of actions. The initial learning rate for both the DNN training and the moments matching was set to $0.01$  with a decaying factor of $1/t$.

\subsection{Catastrophic forgetting} 
First, we examine the impact of catastrophic forgetting on the algorithm's performance on the Statlog dataset. We add a version of the limited memory without moments matching in which the training is done in phases of $400$ steps, similar to the limited-memory variation in \citet{riquelme2018deep}. We ran the experiment with a memory size of $100$ for all limited memory versions to emphasize performance degradation.

Fig~\ref{fig:forgetting} shows the performance of each of the algorithms in this setup. We let each algorithm run for $4000$ steps (contexts) and average each algorithm over $10$ runs. The x-axis corresponds to the number of contexts seen so far, while the y-axis measures the instantaneous regret (lower is better). 
The cumulative reward achieved by each algorithm averaged over seeds can be found in Table~\ref{table:results} (Statlog); in Fig~\ref{fig:forgetting} we focus on the qualitative behavior, as described below. First, we can see that the neural-linear method (blue) outperforms the linear one (brown), suggesting that this dataset requires a nonlinear function approximation. We can also see that LiM2 (orange) perform as well as the neural-linear algorithm without memory constraints (blue) while LiM2 operates online. 

In addition, the limited memory neural-linear algorithms that do not perform likelihood matching (purple and red) suffer from "catastrophic forgetting" due to limited memory. Intuitively, the covariance matrix holds information regarding the number of contexts seen by the agent and used by the algorithm for exploration. When no such prior is available, the agent explores sub-optimal arms from scratch every time the features are modified. In the online version (red), the representation is changed each round, which makes the agent act consistently sub-optimally. Alternatively, in the phase-based version (purple), we see a degradation each time the agent trains (every $400$ steps, marked by the x-ticks on the graph). Indeed,  we observe "peaks" in the regret curve for this algorithm. This is significantly reduced when we compute the prior on the covariance matrix (orange), making the limited memory neural-linear bandit resilient to catastrophic forgetting. The NTK based version (green) acts well but suffers from a "slow" starting due to the representation drift. In the next experiments, we will see that LiM2 performs better overall.  

\begin{figure}[h]
\centering
\centering
 \includegraphics[width=0.8\linewidth]{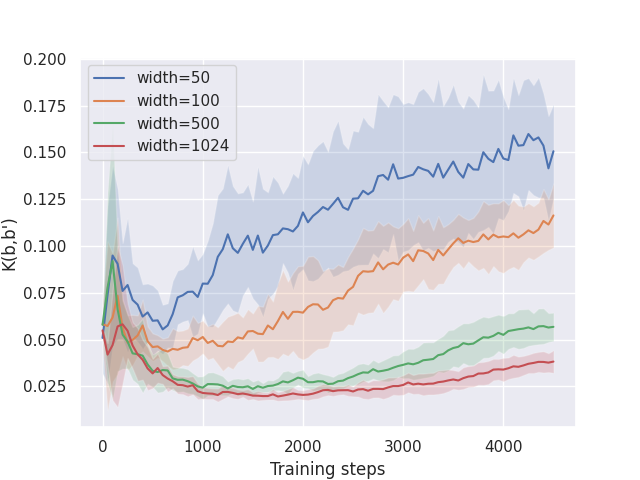}
 \caption{\textbf{Representation drift.} NTK value during training. Representation drift occurs for various network widths}
 \label{fig:Kchange_main}
\end{figure}

\subsection{Representation drift}
Next, we examine the amount of change in the representation $\phi$ during the training of $f_\omega$. To do so, we compute the NTK, denoted by $K$, for two fixed contexts $b$ and $b'$ during training with the Shuttle Statlog dataset \citep{newman2008distributed} for various network widths: $50,100,500,1024$. The results and the experiment for $K(b,b')$
are presented in Fig~\ref{fig:Kchange_main}. More results and details of the experiments are presented in the Supplementary Material at section A. As can be noted, for wider networks, the change at the NTK becomes milder during the training session. Nevertheless, for a significant part of the training, the kernel changes, and the NTK assumption does not hold, which may cause a non-optimal performance. Therefore, a representation drift occurs, and our likelihood matching mechanism is justified. This is of particular importance when exploration and representation learning are done simultaneously as opposed to standard learning problems where no exploration is needed.

\begin{figure}[ht]
\centering
\centering
\includegraphics[width=0.8\linewidth]{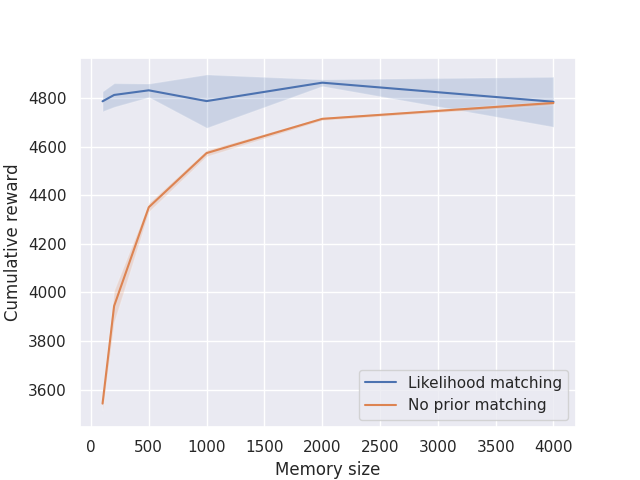}
 \caption{Comparison between our approach with likelihood matching against naive approach without likelihood matching. Our approach is robust to memory size, while naive approach suffers from severe performance degradation with limited memory.}
  \label{fig:mem}
\end{figure}

\subsection{Memory size ablation study} 
We evaluate the impact of the memory size on LiM2. We ran LiM2 with various memory sizes against a limited memory neural-linear without moment matching and checked their performance on the Statlog dataset. For every memory size, we ran the algorithms for $10$ times. The results are depicted in Fig~\ref{fig:mem}. As can be noted, the algorithm performance is not sensitive to the memory size, and its performance across different memory sizes stays high. On the other hand, a version without likelihood matching is deeply sensitive to the memory size due to catastrophic forgetting. Note that the variance margin of the method without likelihood matching is so small compared to the magnitude of the mean, that is not visible in the figure.

\subsection{Real world data} We evaluate our approach on ten real-world datasets and another high-dimensional text dataset (details at the next sub-section); for each dataset, we present the cumulative reward achieved by each algorithm, averaged over $10$ runs. Each run was performed for $5000$ steps. Due to the fact that each dataset behaves differently in terms of difficulty and maximum possible cumulative reward, we normalize the scores for each dataset according to: 
$
normalized\ score(i) = \frac{score(i) - R}{\underset{i}{max}\ score(i) - R},
$
where $R$ is the mean cumulative reward of a random policy. We also present the average and median values for each algorithm. 
The raw results are presented in the Supplementary Material at section B.

Inspecting Table~\ref{table:results}, we can see that using LiM2, improved the performance of the limited memory neural-linear variations, in which we do not update the priors (MM/ Naive). Furthermore, in some cases, LiM2 even outperformed the full memory neural-linear algorithm by a small margin. This can be explained by the fact that limited memory with prior matching behaves as an implicit regularizer by adding noise to the learning. Scania Trucks's results are particularly interesting because the ablative versions (without prior updating) act worse than a random policy, which performs surprisingly well (4663 out of 5000). These findings suggest that likelihood matching improves performance and makes the algorithm resilient to catastrophic forgetting.

In some of the datasets, we observed that a linear baseline (LinearTS) performed well. This suggests that there is no need for a deep representation of the contexts, and that a linear function is good enough to model each hand's reward. A linear algorithm can use finite memory in these cases and does not need to match the likelihood. However, as can be concluded from our results, the reward function is generally non-linear, in which case the linear baseline performs poorly. Nevertheless, even in some datasets where LinearTS performs well, LiM2 outperforms it.

The last part of the table shows the results for the NTK based methods. Clearly, NeuralUCB and NeuralTS are outperformed by LiM2 for most datasets (except Scania Trucks) and poorly perform on the linear datasets. Moreover, NeralLinearTS-NTK shows great performance and outperforms NeuralUCB and NeuralTS, which can be explained by the fact that  NeuralUCB and NeuralTS perform the exploration on the entire network parameter space instead of the last layer as done in the neural-linear case. This makes the problem more complex and harder to solve.
Nevertheless, LiM2 also outperforms NeuralLinearTS-NTK on all datasets. Note that at the Census dataset, a complex dataset in terms of action number and dimensions, the performance gap between the NTK version and LiM2 is significant because the representation problem is harder, and there is a bigger change in the representation during training. These results indicate that our method successfully copes with representation drift.

 \begin{table*}[h]
 \centering
\begin{center}
\tiny
\begin{tabular}{|ccc|c|c|c|c|c|c|c|c|}

\hline
\rule{0pt}{2.6ex} {}    & & & \multicolumn{2}{|c|}{Full memory}  & \multicolumn{3}{c|}{Limited memory} & \multicolumn{3}{c|}{NTK based} \\ \hline

\rule{0pt}{2.6ex} {}  Name & d & A & LinearTS& NeuralLinear&    LiM2 (Ours) & NeuralLinear-MM & NeuralLinear-Naive & NeuralUCB & NeuralTS & NeuralLinear-NTK \\ \hline

\hline
\rule{0pt}{2.6ex} Mushroom & 117  & 2&  \textbf{1.000}    & 0.985        & 0.945           & 0.719  & 0.730    & 0.521     & 0.521    & 0.941  \\ \hline
\rule{0pt}{2.6ex} Financial & 21 & 8 & 0.997    & 0.946        & \textbf{1.000 }          & 0.743  & 0.723    & 0.292     & 0.228    & 0.959 \\ \hline
\rule{0pt}{2.6ex} Jester & 32 & 8 & \textbf{1.000}    & 0.784        & 0.819           & 0.287  & 0.234    & 0.546     & 0.546    & 0.768\\ \hline
\rule{0pt}{2.6ex} Adult & 88 & 2 & 0.977    & 0.974        & \textbf{1.000 }          & 0.638  & 0.634    & 0.822     & 0.823    & 0.966  \\ \hline
\rule{0pt}{2.6ex} Covertype & 54 & 7 & \textbf{1.000}    & 0.902        & 0.892           & 0.679  & 0.693    & 0.514     & 0.517    & 0.887 \\ \hline 

\rule{0pt}{2.6ex} Census & 377 & 9 & 0.548    & 0.860        & \textbf{1.000}           & 0.679  & 0.686    & 0.644     & 0.603    & 0.863  \\\hline 
\rule{0pt}{2.6ex} Statlog & 9 & 7 &0.912    & 0.978        & \textbf{1.000}           & 0.933  & 0.916    & 0.818     & 0.885    & 0.976 \\ \hline
\rule{0pt}{2.6ex} Epileptic  & 178 & 5 &0.282       & \textbf{1.000 }       & 0.684           & 0.562       & 0.504       & 0.019       & 0.020       & 0.589  \\ \hline
\rule{0pt}{2.6ex} Smartphones & 561 & 6 &  0.649    & 0.970        & \textbf{1.000 }          & 0.521  & 0.515    & 0.396     & 0.670    & 0.965\\ \hline
\rule{0pt}{2.6ex} Scania Trucks & 170 & 2 &0.181    & 0.672        & 0.745           & -0.344 & -0.050   & 0.988     & \textbf{1.000 }   & 0.259\\ \hline
\rule{0pt}{2.6ex}{} Amazon & 7K & 5 &-        & 0.986        & \textbf{1.000 }          & 0.873  & 0.879    & -         & -        & 0.981   \\ 

\hline \hline

\rule{0pt}{2.6ex} &  Average & & 0.755  & 0.914  & \textbf{0.917 }  & 0.572  & 0.588 & 0.556    & 0.581& 0.832 \\\hline
\rule{0pt}{2.6ex} {} & Median && 0.945    & 0.970        & \textbf{1.000}           & 0.679  & 0.686    & 0.534     & 0.575    & 0.941 \\\hline
\end{tabular}
\caption{Normalized cumulative reward of algorithms on 11 real world datasets. The context dim $d$ and the size of the action space $A$ are reported for each dataset. The result of each algorithm is reported for $10$ runs. }
\label{table:results}
\end{center}
\end{table*}

\subsection{Sentiment analysis from text using CNNs} This is an experiment on the "Amazon Reviews: Unlocked Mobile Phones" dataset. This dataset contains reviews of unlocked mobile phones sold on "Amazon.com". The goal is to determine the rating (1 to 5 stars) of each review using only the text itself. 
We use our model with a Convolutional Neural Network (CNN) suited to NLP tasks \citep{kim2014convolutional,zahavy2016picture}. Specifically, the architecture is a shallow word-level CNN that was demonstrated to provide state-of-the-art results on various classification tasks by using word embeddings while not being sensitive to hyperparameters \citep{zhang2015sensitivity}. We use the architecture with its default hyper-parameters 
and standard pre-processing (e.g., we use random embeddings of size $128$, and we trim and pad each sentence to a length of 60). The only modification we made was to add a linear layer of size $50$ to make the size of the last hidden layer consistent with previous experiments. The results are in Table~\ref{table:results} marked in yellow.

In this experiment, the input dimension is large ($\mathbb{R}^{7K}$), so we could not run the linear baseline since it is not computationally practical in these dimensions. We compare the proposed method -- neural-linear with finite memory and likelihood matching with the full memory neural-linear TS baseline, limited memory without moment matching baselines, and neural-linear NTK based baseline (although the NTK assumptions do not apply for CNNs).  
Looking at Table~\ref{table:results}, we can see that the limited memory version without likelihood matching performs as good as the full memory and better than the other baselines.

\section{Summary}
\label{sec:dis}
We presented a method for neural-linear contextual bandit with limited memory that succeeds to solve representation drift during exploration, a problem that occurs at NTK based algorithms. Our method is able to do so while being resilient to catastrophic forgetting, which is a major drawback of former limited memory approaches.   
Moreover, we show an efficient way to implement our method by approximately solving an SDP using SGD so that only a single gradient step for both the learning and likelihood matching at each round is sufficient.
Thus, our method is both memory and computationally efficient, which enables it to operate online.     
Our method demonstrated excellent performance on multiple real-world datasets, while its performance did not deteriorate due to the representation changes and limited memory.

We believe that our findings constitute an important step towards solving contextual bandit problems where both exploration and representation learning play essential roles. The main avenue for future work is to extend the ideas presented in this paper to Bayesian RL \cite{adam2020deep}, where the immediate reward is replaced by the return, perhaps focusing on Markov decision processes with fast mixing time (e.g. \cite{zintgraf2019varibad}) or situations in which decisions are infrequent (e.g. decision after a stream of observations) where recurrent models can be used.  

An interesting future work would be to examine the effect of the network architecture on the performance of contextual bandits DNN-based algorithms such as ours and perhaps consider ways to choose the DNN architecture for contextual bandits problems as opposed to this work where the network's architecture is built for supervised learning, which in general may not be optimal for bandit problems.

\section{Acknowledgements}
We would like to thank Nir Levine and David Janz for discussing and providing essential feedback on this work. We would also like to thank Eliav Buchnik for his help and advice in building the experimental setup.

\clearpage

\bibliography{main}
\bibliographystyle{icml2021}

\onecolumn
\icmltitle{Online Limited Memory Neural-Linear Bandits with Likelihood Matching - Supplementary Material}
\section*{A. Representation drift experiment}
The NTK was sampled during training with the Shuttle Statlog dataset \citep{newman2008distributed}.
We compute the NTK for two fixed contexts $b$ and $b'$, taken from the dataset.  Each context is composed of $9$ features describing the space shuttle flight. The goal is to predict the state of the radiator of the shuttle (the reward). There are $N = 7$ possible actions; for correct predictions the reward is $r=1$ and $r=0$ otherwise.

The weights of the network were initialized according to \cite{jacot2018neural}. We ran the experiments for various network widths: 50, 100, 500, and 1024. The network was trained as described in the main paper (Method and setup) with a random policy. The NTK was sampled every $50$ iterations during training for the estimated reward of the first action. The graphs below present $K(b,b), K(b',b')$ and $K(b,b')$ from left to right accordingly during training averaged over $10$ seeds.

\begin{figure}[h]
\centering
\begin{minipage}{.3\textwidth}
  \centering
  \includegraphics[width=1.1\linewidth]{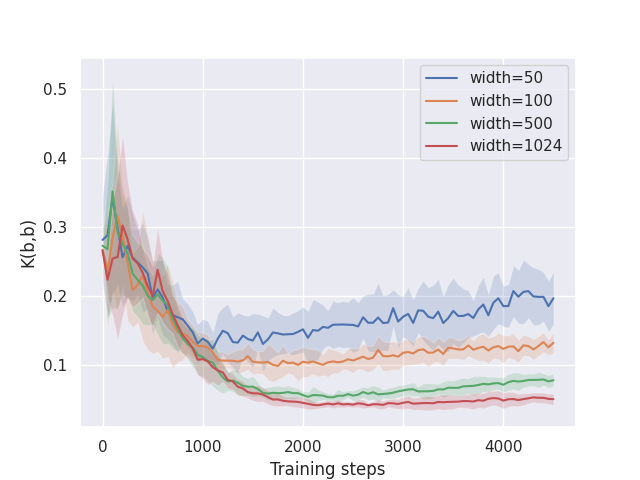}
\end{minipage}%
\begin{minipage}{.3\textwidth}
  \centering
  \includegraphics[width=1.1\linewidth]{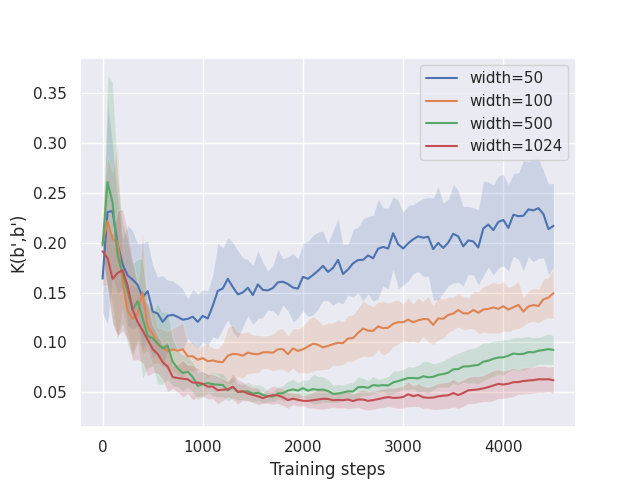}
\end{minipage}
\begin{minipage}{.3\textwidth}
  \centering
  \includegraphics[width=1.1\linewidth]{figures/Kfig31.png}
\end{minipage}
\label{fig:Kchange}
\end{figure}


\section*{B. Raw results}
\label{sec:rawresults}

\begin{table}[h]
\label{table:rawresults}
\begin{center}
\tiny
\begin{tabular}{|ccc|c|c|c|c|c|c|c|c|}
\hline
\rule{0pt}{2.6ex}{}    & & & \multicolumn{2}{|c|}{Full memory}  & \multicolumn{3}{c|}{Limited memory} & \multicolumn{3}{c|}{NTK based} \\ \hline
\rule{0pt}{2.6ex} {}  Name & d & A & LinearTS& NeuralLinear&    LiM2 (Ours) & NeuralLinear-MM & NeuralLinear-Naive & NeuralUCB & NeuralTS & NeuralLinear-NTK  \\ \hline

\hline
\rule{0pt}{2.6ex} Mushroom & 117  & 2&  \textbf{ 11162 $\pm$ 1167} & 10810 $\pm$ 428  &  9880 $\pm$ 1776  &  4602 $\pm$ 1408 &4843 $\pm$ 1228& -32 $\pm$ 3 & -34 $\pm$ 18 & 9785 $\pm$ 1012 \\ \hline
\rule{0pt}{2.6ex} Financial & 21 & 8 & 3752 $\pm$ 8 & 3560 $\pm$ 12& \textbf{3762 $\pm$ 18} &2802 $\pm$ 21 & 2726 $\pm$ 23& 1116 $\pm$ 824& 876 $\pm$ 595& 3610 $\pm$ 18\\ \hline
\rule{0pt}{2.6ex} Jester & 32 & 8 & \textbf{15944 $\pm$ 170} & 14731 $\pm$ 304 &  14926 $\pm$ 571 & 11940 $\pm$ 1307 &11647 $\pm$ 1066 &13397 $\pm$ 7& 13397 $\pm$ 8 & 14642 $\pm$ 200\\ \hline
\rule{0pt}{2.6ex} Adult & 88 & 2 & 4008 $\pm$ 14   & 4003 $\pm$ 12  &  \textbf{4043 $\pm$ 15} & 3483 $\pm$ 33 & 3477 $\pm$ 16.5&3768 $\pm$ 2 &3769 $\pm$ 2 & 3990 $\pm$ 17 \\ \hline
\rule{0pt}{2.6ex} Covertype & 54 & 7 &\textbf{2961 $\pm$ 25} & 2742 $\pm$ 40 &   2719 $\pm$ 62 & 2241 $\pm$ 30 & 2272 $\pm$ 37 &1870 $\pm$ 11 & 1877 $\pm$ 83 & 2708 $\pm$ 31\\ \hline 

\rule{0pt}{2.6ex} Census & 377 & 9 & 1801 $\pm$ 15 & 2510 $\pm$ 21& \textbf{2827 $\pm$ 22}& 2099 $\pm$ 39& 2114 $\pm$ 34& 2019 $\pm$ 94 & 1926 $\pm$ 76& 2517 $\pm$ 42\\\hline 
\rule{0pt}{2.6ex} Statlog & 9 & 7 &4460 $\pm$ 19 & 4729 $\pm$ 6 &  \textbf{4820 $\pm$ 68}& 4545 $\pm$ 34 &4476 $\pm$ 19 &4075 $\pm$ 3 &4348 $\pm$ 265 & 4722 $\pm$ 12\\ \hline
\rule{0pt}{2.6ex} Epileptic  & 178 & 5 &1204 $\pm$ 29 &\textbf{1734 $\pm$ 46} &1501 $\pm$ 115 & 1411 $\pm$ 30& 1368 $\pm$ 26 &1010 $\pm$ 2 & 1011 $\pm$ 2 &1431 $\pm$ 47  \\ \hline
\rule{0pt}{2.6ex} Smartphones & 561 & 6 &  3092 $\pm$ 18 &4208 $\pm$ 23 &\textbf{4313 $\pm$ 46}& 2647 $\pm$ 17& 2626 $\pm$ 16& 2214 $\pm$ 1548 &3166 $\pm$ 1113 & 4191 $\pm$ 32\\ \hline
\rule{0pt}{2.6ex} Scania Trucks & 170 & 2 &4710 $\pm$ 171 & 4837 $\pm$ 132 & 4856 $\pm$ 111  &4574 $\pm$ 220   &4650 $\pm$ 103   &4919 $\pm$ 19 &\textbf{4922 $\pm$ 23}  &  4730 $\pm$  179\\ \hline
\rule{0pt}{2.6ex} Amazon & 7K & 5 & - & 3024 $\pm$ 25 & \textbf{3052 $\pm$ 160}  &2793 $\pm$ 41& 2804 $\pm$ 41   & - & -  &  3014 $\pm$  37\\ \hline
\end{tabular}
\end{center}
\caption{Cumulative reward on 11 real world datasets.}
\end{table} 

\end{document}